\documentclass{article}

\PassOptionsToPackage{numbers, compress}{natbib}
\usepackage[final]{nips_2017}

\usepackage[utf8]{inputenc} 
\usepackage[T1]{fontenc}    
\usepackage{hyperref}       
\usepackage{url}            
\usepackage{booktabs}       
\usepackage{amsfonts}       
\usepackage{nicefrac}       
\usepackage{microtype}      
\usepackage{graphicx}
\usepackage{placeins}
\usepackage{svg}
\usepackage{caption}

\title{Anesthesiologist-level forecasting of hypoxemia with only SpO\textsubscript{2} data using deep learning}

\author{
  Gabriel Erion\\
  University of Washington\\
  \texttt{erion@cs.washington.edu} \\
  \\
  \And
  Hugh Chen \\
  University of Washington\\
  \texttt{hughchen@uw.edu} \\
  \And
  Scott M. Lundberg \\
  University of Washington\\
  \texttt{slund1@cs.washington.edu} \\
  \And
  Su-In Lee\\
  University of Washington\\
  \texttt{suinlee@cs.washington.edu} \\
}

\begin{document}

\maketitle

\begin{abstract}
 We use a deep learning model trained only on a patient's blood oxygenation data (measurable with an inexpensive fingertip sensor) to predict impending hypoxemia (low blood oxygen) more accurately than trained anesthesiologists with access to all the data recorded in a modern operating room. We also provide a simple way to visualize the reason why a patient's risk is low or high by assigning weight to the patient's past blood oxygen values. This work has the potential to provide cutting-edge clinical decision support in low-resource settings, where rates of surgical complication and death are substantially greater than in high-resource areas.
\end{abstract}

\section{Introduction}

Over 5 billion patients worldwide lack safe, affordable access to necessary surgical and anesthetic care
\citep{meara2015global}.
While anesthesia-related mortality has declined substantially since 1940 in high-Human Development Index (HDI) countries, it has actually increased in low-HDI countries. A recent meta-analysis showed two to three times greater anesthesia-related mortality in low-HDI countries, but such mortality in some major hospitals in Sub-Saharan Africa is up to 400 times greater.
\citep{pollach2013anaesthetic,walker2008anaesthesia,cherian2010building}

An important component of mortality and morbidity surrounding a surgery (\textit{perioperative} mortality and morbidity) is hypoxemia, or low oxygen levels in the blood. Perioperative hypoxemia is common; prolonged hypoxemia can cause serious adverse cardiac and neurological effects.
\citep{strachan2001hypoxia}
The risks posed by hypoxemia may be one reason that pulse oximeters (devices that measure blood oxygen) are promoted by nonprofit organizations such as \citep{enright2016lifebox} as a way to improve surgical safety in low-HDI countries. 
Shortages of essential equipment (such as pulse oximeters) are indeed a major factor driving differences in surgical risk between low- and high-HDI countries. However, a lack of trained anesthesiologists is also a major cause (one study found 96 percent of anesthesia providers in Uganda are non-physicians) that equipment alone cannot fix \citep{cherian2010building}.
This lack of training may be particularly important because delay in recognition and treatment of adverse events during surgery is an important component of perioperative mortality and morbidity \citep{stiegler2014cognitive}.

Our paper addresses the dangers of delay in hypoxemia recognition by developing a system which can predict hypoxemia as well as or better than physician anesthesiologists, using pulse oximetry data alone. We believe this system can, at no additional cost, enable pulse oximeters currently being distributed in low-resource settings to provide advance warning of hypoxemia.

\textbf{Related Work}
Machine learning for clinical prediction is growing more popular, and several recent methods have achieved doctor-level prediction performance on certain tasks.
Two recent papers showed that Inception-based convolutional neural networks performed as accurately as physicians in diagnosing images of possible skin cancer and diabetic retinopathy \citep{esteva2017dermatologist, gulshan2016development}.
In non-image prediction, another paper successfully classified heart arrhythmias as accurately as cardiologists by doing 1-dimensional convolutions over time-series electrocardiogram recordings \citep{rajpurkar2017cardiologist}.
Finally,
\citep{lundberg2017explainable}
used all data gathered in the operating room during tens of thousands of surgeries to train a gradient boosting classifier that predicted perioperative hypoxemia more accurately than anesthesiologists. Inspired by this clinical task, and by the fact that \citep{lundberg2017explainable}'s feature importance method assigned by far the most importance to patient SpO$_2$ (blood oxygen percentage), this paper attempts to outperform anesthesiologists using a subset of the same data consisting \textit{only of SpO$_2$ values.}

\section{Operating-Room Hypoxemia Prediction}
\subsection{Data}
Our data came from an academic medical center's Anesthesia Information Management System (AIMS), which records all data measured in the operating room in real time during surgery. This includes demographic data (age, sex, height, weight), diagnosis and procedure codes,  and free text in the medical record, as well as real-time measurements of vital signs, laboratory results, and drug doses as they are given. We trained on data from 57,173 cases, split into 8,088,523 training time points, and tuned our models using a 90-10 train-validation split. Time points from all cases were shuffled and used as independent datums; investigating per-patient effects and dependency between time points could be a useful future goal. We compared all models' performances on a set of 1,053,116 held-out testing time points from 7,569 separate cases. Finally, we compared the best-performing model against anesthesiologists' predictions from a user study on an entirely separate 523 time points.

While other work with this system has used all available AIMS data, we limited our data at each time point to 60 SpO$_2$ measurements, one per minute for the past hour of surgery. 
We imputed missing values and normalized each column to 0 mean and unit variance. The label at each time point was zero if hypoxemia, measured as a drop in SpO$_2$ to 92 percent or lower, did not occur in the next five minutes. If SpO$_2$ did drop to 92 percent or lower, the label was one. Points where SpO$_2$ has dropped below 95 percent in the past ten minutes were not considered.
In the 523 anesthesiologist comparison time points, cases where SpO$_2$ did not drop to 92 percent or lower in the next five minutes but did in the next ten were excluded, so that doctors were only tested on clearly positive or negative cases. This shift in data distribution may explain the difference in ROC between LSTMs in the model comparison (Figure 1, left) and anesthesiologist performance comparison (Figure 1, right).

\subsection{Models}
We considered three complex models for hypoxemia prediction, as well as base rate, ARIMA(1,0,0) and logistic regression predictors (Figure 1, left), and compared the best-performing model against doctors (Figure 1, right).  Because the base rate of hypoxemia in this data was very low (1.7 percent), we decided to compare models using area under the precision-recall curve (AU-PRC), which best distinguishes model performance with imbalanced classes. We trained the gradient boosted model on the full data as one batch, but fed data to the neural nets in balanced batches to improve convergence.

\textbf{Gradient boosted trees}
As in \citep{lundberg2017explainable}
we used the fast XGBoost implementation of gradient boosting.
\citep{chen2016xgboost}
In tuning on the validation set, we found that, with a learning rate of $\eta=0.01$, using 4400 trees of depth 6 resulted in the best performance. Turning $\eta$ from 0.1 to 0.01 resulted in longer training time but improved performance. The AU-PRC on the test data was 0.22642.

\textbf{Convolutional neural network}
We also considered a 1-dimensional convolutional network modeled on \citep{rajpurkar2017cardiologist}. The first convolutional layer consisted of a convolution followed by batch normalization and a ReLU activation. Five more layers were added, with the structure (batch normalization, ReLU, dropout, convolution). The final layer had the structure (batch normalization, ReLU, dense, sigmoid). Every convolution had kernel size 6. The first two convolutions used 64 filters; all subsequent ones used 128. Unlike in \citep{rajpurkar2017cardiologist}
residual connections did not improve performance on the validation data -- likely due to the shallow network -- and were not used. A shallower net seems justified as \citep{rajpurkar2017cardiologist} used 6000 measurements per time point while we have only 60.
The network was trained with the Adam optimizer \cite{kingma2014adam}. Final AU-PRC on the test data was 0.22202, less than the XGBoost and LSTM models and somewhat surprising given the recent popularity of convolutional networks for time series classification. It is possible that a deeper network or more parameter tuning was needed.

\textbf{Long short-term memory network}
Our final model was a long short-term memory network (LSTM), whose recurrent structure is a natural fit for time series data. The LSTM consisted of a 200-node LSTM layer with recurrent dropout on top of the input data; this layer output a sequence which was fed to another 200-node LSTM layer with recurrent dropout. This layer produced a single output, which was fed through a single dense node to a sigmoid output. There was dropout between all layers, and the network was optimized with Rmsprop \cite{tieleman2012lecture}. The final AU-PRC on test data was 0.23142.

\begin{figure}[h]
  \begin{minipage}[b]{0.50\linewidth}
  {\caption*{\textbf{Model comparison and selection}}}
  \centering
    \begin{tabular}{ |c|l|l| } 
     \hline
     Model & AU-PRC & AU-ROC \\
     \hline 
     Base Rate & 0.017313 & 0.49938 \\ 
     ARIMA(1,0,0) & 0.022153 & 0.50529 \\
     Logistic Regression & 0.12918 & 0.74703 \\ 
     Convolutional net & 0.22202 & 0.86134 \\ 
     Gradient boosting & 0.22641 & 0.86363 \\ 
     \textbf{LSTM} & \textbf{0.23139} & \textbf{0.86571} \\ 
     \hline
    \end{tabular}
    \par\vspace{30pt}
    \end{minipage}
  \begin{minipage}[b]{0.50\linewidth}
  \begin{center}
  \caption*{\textbf{Anesthesiologist performance comparison}}
  \end{center}
  \centering
  \includegraphics[width=0.75\linewidth]{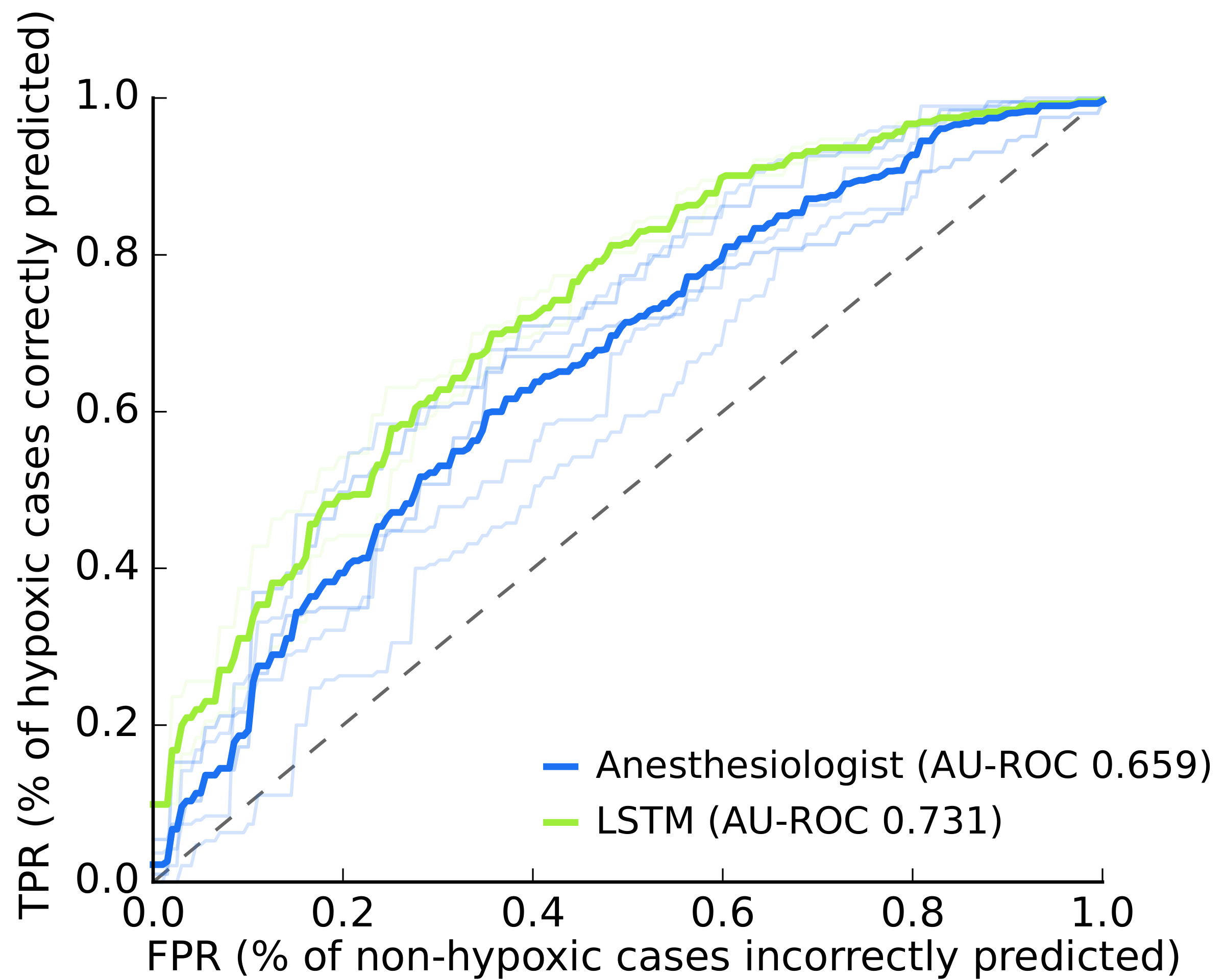}
  \par\vspace{0pt}
  \end{minipage}
  \caption{Left: Performance of all models on 1,053,116 test points, measured as area under precision-recall and ROC curves.
  Right: Comparison of deep learning LSTM model and anesthesiologist predictions of hypoxemia on 523 time points. LSTM was given access only to 60 minutes of oxygen data, while anesthesiologists were given access to all data recorded in the OR as plots and notes in a web interface. The performance difference is significant with P<0.0001.}
\end{figure}

\subsection{Comparison with Doctors}
We used the receiver operator characteristic (ROC) curve, a widespread standard for medical diagnostics, to compare the best-performing model, the LSTM, against anesthesiologists. Because a \textit{precision-recall} curve had been used to choose the best-performing model, we verified that the ranking of models (LSTM>XGBoost>CNN) was the same under both AU-PRC and AU-ROC (Figure 1, left). The LSTM model had noticeably better AU-ROC (0.731) than doctors' pooled AU-ROC of 0.659, with P<.0001 calculated by bootstrap (Figure 1, right). The LSTM curve also dominates the doctors' curve; for every possible given false positive rate, it achieves a higher true positive rate.

\subsection{Model Interpretation}
We agree with the observation in \citep{lundberg2017explainable}
that explaining the predictions of a clinical model is essential for doctors to trust and use the model. Unlike in \citep{lundberg2017explainable},
our model only uses 60 sequential values per prediction, so it is actually possible to visualize the entire data input and the importance of each feature. Recent methods have been developed that estimate feature importances for individual predictions in each of the models we use: Tree SHAP for XGBoost \citep{lundberg2017consistent}
and Integrated Gradients for LSTM/CNN \citep{sundararajan2017axiomatic,hiranuma2017integrated}.
In Figure 2, we show both the data for a single case and the feature importances for each model at each time point. In general, the models behave as one would predict; most features have little contribution to risk but drops in oxygen have a large positive one. The closer a drop in oxygen is to the present, the larger its contribution to the risk. The convolutional network sometimes exhibits confusing behavior, handling slow increases in SpO$_2$ by creating large periodic waves of risk contribution (see row 3, columns 1 and 2). This could be due to the fixed-size window of the convolutions trying to average out to a long term trend, or it could be due to instability from the fact that the convolutional model is the deepest. Some of this effect can occasionally be seen in the LSTM (row 4, column 2), though it appears overall more stable.

\begin{figure}[h]
  \caption*{\textbf{Model explanations}}
  \vspace{-5pt}
  \centering
  \includegraphics[width=\linewidth]{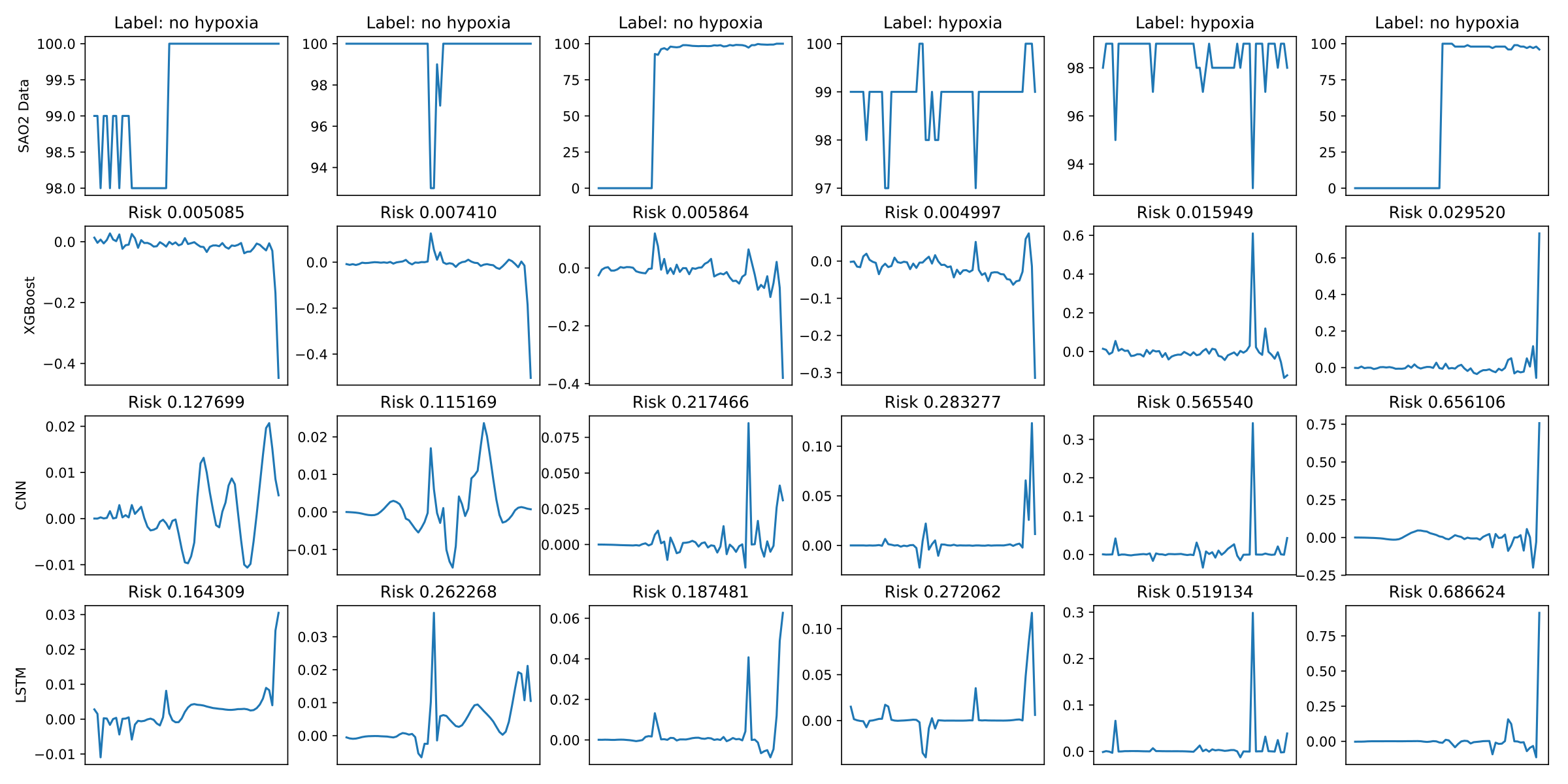}
  \caption{Model explanations for all models. Each column is 60 minutes of SpO$_2$ data from a test time point shown to doctors (60 minutes ago at left, current time at right). Top row is observed SpO$_2$ data over 60 minutes. Subsequent rows are the importance of each minute's SpO$_2$ value for the XGBoost, CNN, and LSTM models. Predicted risk is shown above each model's explanation plot. Note that XGBoost predicts lower risks than neural nets, as it was not fed data in balanced batches.}
\vspace{-15pt}
\end{figure}

\section{Discussion}
We have presented a method that builds on previous work by giving advance warning of hypoxemic events using only easily-measured SpO$_2$ data. The system outperforms anesthesiologists who have access to all the data recorded in a modern operating room (SpO$_2$ plus other vital signs, demographics, medical record, drugs given, etc). Because the user study was done on computers, anesthesiologists did not have physical access to the patient; however, they still enjoyed a substantial data advantage over our model. The model's success raises the interesting question of whether doctors might have been better able to focus on the most important variable and make more accurate predictions if given a more limited dataset -- for example, just SpO$_2$. 
It is also worth noting that anesthesiologists do not train to be experts in predicting hypoxia but in clinical management of an unconscious, non-breathing patient (while radiologists, a common comparison for machine learning methods, do indeed train to be experts in diagnosis based on images). This does not detract from the clinical value of our problem; in fact, algorithms to supplement clinician judgment on ancillary tasks like hypoxemia prediction may free up attention for the many other jobs an anesthesiologist must do during surgery.

Another important contribution of this work is demonstrating that inexpensive sensors like pulse oximeters can have great predictive power in the operating room (building on \citep{rajpurkar2017cardiologist}'s use of a single-lead EKG sensor to classify arrhythmias). This implies two future research directions: First, other perioperative adverse events may be predictable with other simple sensors. Hypocapnia (low expired CO$_2$, measured with an end-tidal CO$_2$ sensor) is associated with longer ICU stays after surgery, while hypotension (low blood pressure, measured with a non-invasive blood pressure cuff) is associated with increased mortality \citep{jeremitsky2003harbingers}. Predictive models for these end points would likely have substantial clinical value.
Second, waveform sensors like oximeters (and ETCO$_2$ capnography) record signals at 100Hz or greater; AIMS stores a far lower-resolution signal. Access to the raw waveform signal would augment our data by a factor of 6000 and almost certainly lead to more accurate predictions.

Finally, this work has important implications for clinical decision support in low-resource settings. While an AIMS system costs tens to hundreds of thousands of dollars \citep{ehrenfeld2011anesthesia},
pulse oximeters distributed by nonprofit organizations cost as little as \$250 \citep{enright2016lifebox} and may often be the only monitoring equipment available. Algorithms like ours have the potential to, at no extra cost, turn these oximeters into a source of decision support as effective (or more so) than a trained physician in anticipating adverse events. We hope that such a tool would contribute to reducing the dramatic disparity in perioperative risk faced in low-resource settings and helping to make surgery safer worldwide.

\FloatBarrier



\medskip
  
\bibliography{nips_2017.bib}

\small

\end{document}